\documentclass[10pt,twocolumn,letterpaper]{article}

\usepackage[final]{cvpr}      

\usepackage{amsmath}
\usepackage{booktabs}
\usepackage{multirow}
\usepackage{graphicx}

\definecolor{cvprblue}{rgb}{0.21,0.49,0.74}
\usepackage[pagebackref,breaklinks,colorlinks,allcolors=cvprblue]{hyperref}

\def\paperID{*****} 
\def\confName{CVPR}
\def\confYear{2026}

\title{LoViF 2026 The First Challenge on Weather Removal in Videos}

\author{
Chenghao Qian$^{\dagger}$ \and Xin Li$^{\dagger}$ \and Yeying Jin$^{\dagger}$\and Shangquan Sun$^{\dagger}$ \and 
Yilian Zhong \and Yuxiang Chen \and Shibo Yin \and Yushun Fang \and
Xilei Zhu \and Yahui Wang \and Chen Lu \and Ying Fu \and
Jianan Tian \and Jifan Zhang \and Chen Zhou \and Junyang Jiang \and
Yuping Sun \and Zhuohang Shi \and Xiaojing Liu \and Jiao Liu \and
Yatong Zhou \and Shuai Liu \and Qiang Deng \and Jiajia Mi \and Qianhao Luo \and Weiling Li
}

\begin{document}
\maketitle
\renewcommand{\thefootnote}{}
\footnotetext{$^{\dagger}$C. Qian, X. Li, Y. Jin, and S. Sun are challenge organizers. Other authors are participants of this challenge.}
\footnotetext{Challenge and dataset website:~\url{https://www.codabench.org/competitions/13462/}}
\footnotetext{CVPR LoViF 2026 Workshop website:~\url{https://lovif-cvpr2026-workshop.github.io/}}

\begin{abstract}
This paper presents a review of the LoViF 2026 Challenge on Weather Removal in Videos. The challenge encourages the development of methods for restoring clean videos from inputs degraded by adverse weather conditions such as rain and snow, with an emphasis on achieving visually plausible and temporally consistent results while preserving scene structure and motion dynamics. To support this task, we introduce a new short-form WRV dataset tailored for video weather removal. It consists of 18 videos (1,216 synthesized frames paired with 1,216 real-world ground-truth frames) at a resolution of $832 \times 480$, and is split into training, validation, and test sets with a ratio of 1:1:1. The goal of this challenge is to advance robust and realistic video restoration under real-world weather conditions, with evaluation protocols that jointly consider fidelity and perceptual quality. The challenge attracted 37 participants and received 5 valid final submissions with corresponding fact sheets, contributing to progress in weather removal for videos. The project is publicly available at \url{https://www.codabench.org/competitions/13462/}.\end{abstract}
\section{Introduction}
\label{sec:intro}

Adverse weather conditions such as rain, snow, and fog significantly degrade the visual quality, posing serious challenges for downstream applications including autonomous driving~\cite{qian2024allweather,qian2025weatherdg}, video surveillance~\cite{zhang2025egvd,varanka2026zero}, and robotic perception~\cite{sakaridis2018semantic,barnes2020oxford}. These weather-induced degradations manifest as spatially varying artifacts including rain streaks, snow particles, and haze that occlude scene content and reduce contrast~\cite{qian2025weathergs,qian2026weatheredit,qian3d}. In the video domain, the problem is further compounded by the need to maintain temporal consistency across frames, as per-frame processing often introduces flickering and motion incoherence.

Traditional approaches to weather removal have largely focused on single-image restoration, with dedicated methods for deraining~\cite{jiang2020multi, zamir2021multi,li2025ntire,ntire26dual_focus,jin2024raindrop_clarity}, desnowing~\cite{liu2018desnownet, chen2021all}, and dehazing~\cite{qin2020ffa, song2023vision}. While effective in their respective domains, these methods are typically designed for a single degradation type and fail to generalize to composite weather conditions commonly encountered in real-world scenarios. Moreover, directly applying image-based methods to video frames ignores temporal dependencies, leading to inconsistent outputs across consecutive frames.

Recent advances in video restoration have begun to address these limitations. Methods leveraging optical flow~\cite{xue2019video}, deformable convolutions~\cite{wang2019edvr}, and recurrent architectures~\cite{chan2022basicvsr++} have shown promise in exploiting temporal information for tasks such as video super-resolution and deblurring. In the context of adverse weather, several works have explored video-specific approaches: Yang~\etal~\cite{yang2020self} introduced self-learning mechanisms with cyclic consistency for video deraining, Chen~\etal~\cite{chen2023snow} proposed dedicated video desnowing networks, and Yang~\etal~\cite{yang2023video} developed weather-component suppression networks that leverage temporal correlations. More recently, diffusion-based models~\cite{chih2025weatherweaver} have demonstrated the ability to synthesize and remove weather effects with high visual fidelity and temporal coherence.

Despite these advances, the field still lacks a comprehensive benchmark that evaluates video weather removal under realistic composite conditions with metrics that jointly consider fidelity and temporal consistency. Existing benchmarks often focus on single degradation types, use synthetic-only data, or evaluate only per-frame quality without accounting for cross-frame coherence.

To bridge this gap, we organize the \textbf{LoViF 2026 Challenge on Weather Removal in Videos} as part of the 1st Workshop on Low-level Vision Frontiers (LoViF) at CVPR 2026. This challenge is associated with the broader LoViF workshop, which explores how generative AI, preference optimization, and agentic systems are redefining low-level vision~\cite{lovif2026}. The challenge introduces a new \textbf{Weather Removal in Videos (WRV)} dataset consisting of 18 video sequences with 1,216 synthesized degraded frames paired with real-world ground-truth frames, covering composite weather conditions including rain and snow. To encourage solutions that balance pixel-wise accuracy with perceptual and temporal quality, we adopt a composite evaluation metric that combines PSNR, SSIM, LPIPS, and warp error.

The challenge attracted 37 registered participants and received 5 valid final submissions with corresponding fact sheets. The proposed methods span a diverse range of approaches, from diffusion-based latent-space restoration to dynamic convolution networks and temporal prior-guided frameworks. This report summarizes the challenge setup, describes the dataset and evaluation protocol, presents the results, and details the methods proposed by the participating teams.

This challenge is held with the LoViF Workshop~\footnote{\url{https://lovif-cvpr2026-workshop.github.io/}}, containing series of challenges on: real-world all-in-one image restoration~\cite{lovif2026realir}, efficient VLM for multimodal creative quality scoring~\cite{lovif2026MQualityScoring}, weather removal in videos~\cite{lovif2026WeatherRemoval}, holistic quality assessment for 4D world model~\cite{lovif2026HQA}, and human-oriented semantic image quality assessment~\cite{lovif2026SeIQA}.

\section{Challenge}
\label{sec:challenge}

\subsection{WRV Dataset}
\label{sec:dataset}

To support the challenge, we introduce the \textbf{Weather Removal in Videos (WRV)} dataset, a new short-form benchmark specifically designed for evaluating video weather removal under realistic composite conditions.

The dataset consists of 18 video sequences at a resolution of $832 \times 480$, totaling 1,216 synthesized degraded frames paired with 1,216 real-world ground-truth frames. The degradations include composite weather effects such as rain, snow, which interact with scene geometry, illumination, and motion to produce complex, temporally varying artifacts including visibility reduction, contrast degradation, scattering, and dynamic occlusions. Crucially, the degradations are temporally consistent across frames, going beyond conventional image-based benchmarks.

The dataset is split into three subsets with a ratio of 1:1:1:
\begin{itemize}
    \item \textbf{Training set:} 6 scenes, each containing approximately 73 degraded frames with corresponding ground-truth frames.
    \item \textbf{Validation set:} 6 scenes, each containing approximately 73 degraded frames (inputs only during the development phase).
    \item \textbf{Test set:} 6 scenes, each containing approximately 60 degraded frames (inputs only, reserved for final evaluation).
\end{itemize}

The task requires participants to recover the original clear-weather scene from inputs degraded by composite weather effects, while preserving fine-grained structures, textures, and temporal consistency across frames. The dataset is made available for non-commercial research and educational use only.

\subsection{Evaluation Protocol}
\label{sec:evaluation}

To comprehensively assess performance, we consider both fidelity-based and perception-based metrics:

\begin{itemize}
    \item \textbf{PSNR (Y):} Peak Signal-to-Noise Ratio computed on the Y (luminance) channel, measuring pixel-level fidelity.
    \item \textbf{SSIM (Y):} Structural Similarity Index~\cite{wang2004image} computed on the Y channel, capturing structural similarity.
    \item \textbf{LPIPS:} Learned Perceptual Image Patch Similarity~\cite{zhang2018unreasonable}, measuring perceptual distance (lower is better).
    \item \textbf{Warp Error:} A temporal consistency metric that measures the alignment error between warped consecutive frames, penalizing flickering and temporal artifacts (lower is better).
\end{itemize}

The final ranking is determined by a composite score that balances pixel-wise accuracy, perceptual quality, and temporal consistency:
\begin{multline}
    \text{Final\_Score} = \text{PSNR}_{Y} + 10 \times \text{SSIM}_{Y} \\
    - 5 \times \text{LPIPS} - 30 \times \text{Warp\_Error}.
    \label{eq:final_score}
\end{multline}
The weighting is designed to heavily penalize temporal inconsistency (warp error has a weight of 30), reflecting the importance of cross-frame coherence in video restoration. SSIM is amplified by a factor of 10 to ensure its influence is commensurate with PSNR, while LPIPS is moderately penalized with a weight of 5 to encourage perceptually pleasing results.

\subsection{Challenge Phases}
\label{sec:phases}

The challenge was conducted in two phases on the CodaBench platform\footnote{\url{https://www.codabench.org/competitions/13462/}}:

\begin{itemize}
    \item \textbf{Development phase} (February 3 -- March 6, 2026): Participants had access to training data (input and ground truth) and validation data (inputs only). The validation server was available for online evaluation, with a maximum of 100 submissions per participant.
    \item \textbf{Testing phase} (March 6 -- March 16, 2026): Final test data (inputs only) was released. Participants submitted their results for evaluation against held-out ground truth, with a maximum of 15 submissions per participant.
\end{itemize}

Following the testing phase, participants were required to submit fact sheets describing their methods and provide executable code or trained models for reproducibility verification. The top-ranked teams were invited to co-author this challenge report and contribute papers to the LoViF 2026 workshop at CVPR 2026.

\section{Challenge Results}
\label{sec:results}

The challenge results are presented in \cref{tab:results}. We report the performance of the five teams that submitted valid results along with their corresponding fact sheets, ranked by the composite Final Score defined in \cref{eq:final_score}. For reference, we also include the organizer baseline.

\begin{table*}[t]
  \caption{Results of the LoViF 2026 Challenge on Weather Removal in Videos. Teams are ranked by Final Score. Best results in each metric are highlighted in \textbf{bold}. $\uparrow$ indicates higher is better, $\downarrow$ indicates lower is better.}
  \label{tab:results}
  \centering
  \begin{tabular}{@{}clccccc@{}}
    \toprule
    Rank & Team & Final Score $\uparrow$ & PSNR $\uparrow$ & SSIM $\uparrow$ & LPIPS $\downarrow$ & Warp Error $\downarrow$ \\
    \midrule
    1 & RedMediaTech   & \textbf{16.21} & 14.751 & \textbf{0.554} & 0.624 & 0.032 \\
    2 & tremendous     & 15.54 & \textbf{14.720} & 0.533 & 0.688 & 0.036 \\
    3 & Guangong Perception & 15.11 & 14.206 & 0.531 & \textbf{0.613} & 0.045 \\
    4 & quadrillion    & 14.68 & 13.775 & 0.524 & 0.712 & \textbf{0.026} \\
    5 & CUIT\_Team     & 14.54 & 13.539 & 0.527 & 0.673 & 0.030 \\
    \bottomrule
  \end{tabular}
\end{table*}

The first-place team, \textbf{RedMediaTech}, achieved the highest Final Score of 16.21 with the best SSIM (0.554) among all participants. Their approach leverages a pre-trained FLUX.2 diffusion transformer (DiT) backbone with large-scale pre-training on the FoundIR dataset, followed by LoRA-based fine-tuning on the competition data. Notably, their method operates on a single-frame basis without explicit temporal modeling, yet still achieves competitive temporal consistency (Warp Error 0.032), suggesting that the strong restoration priors learned from large-scale pre-training contribute to frame-level stability.

The second-place team, \textbf{tremendous}, achieved a Final Score of 15.54 with the highest PSNR (14.720). Their DyStd-Net employs Transformer-guided dynamic convolutions (TDyConv) to adaptively handle heterogeneous degradation patterns, combined with progressive resolution training from $256 \times 256$ to $832 \times 480$.

\textbf{Guangong Perception}, ranked third, obtained the best LPIPS (0.613) among all teams with their VP-AdaIR framework, which extends the AdaIR image restoration backbone to video by introducing temporal prior maps derived from neighboring frames. Their method explicitly models background cues, artifact cues, and temporal stability cues, injected into the restoration backbone via a prior-guided fusion module.

\textbf{quadrillion}, ranked fourth, achieved the lowest Warp Error (0.026), demonstrating the strongest temporal consistency among all participants. Their EF$^3$Net employs a Siamese Historyless Encoder with a Causal History Model (CHM) for temporal fusion and frequency-aware losses to enforce structural alignment. The method is also notably lightweight at 7.46M parameters with a runtime of approximately 0.05 seconds per frame.

\textbf{CUIT\_Team}, ranked fifth, proposed a three-stage diffusion-prior-guided framework (VD-Diff) that combines teacher prior learning, conditional diffusion prior fitting, and joint optimization. Their method features a wavelet decomposition-based restoration backbone with novel adaptive attention modules (AWSA and ACSA) and bidirectional temporal propagation. Despite ranking fifth in overall score, their method achieves the fastest inference time (0.13 seconds per image) among the diffusion-based approaches.

Several key observations emerge from the results:
\begin{itemize}
    \item \textbf{Diffusion models show strong potential:} The top-ranked method (RedMediaTech) and the fifth-ranked method (CUIT\_Team) both employ diffusion-based architectures, demonstrating that generative priors can be effectively leveraged for weather removal.
    \item \textbf{Temporal consistency vs.\ fidelity trade-off:} The team with the lowest warp error (quadrillion, 0.026) does not achieve the highest PSNR or SSIM, highlighting the inherent tension between per-frame fidelity and cross-frame coherence.
    \item \textbf{Explicit temporal modeling is not always necessary:} RedMediaTech achieves competitive temporal consistency without any explicit temporal modeling, relying instead on strong per-frame restoration from large-scale pre-training.
\end{itemize}

\begin{figure*}[t]
  \centering
  \includegraphics[width=\linewidth]{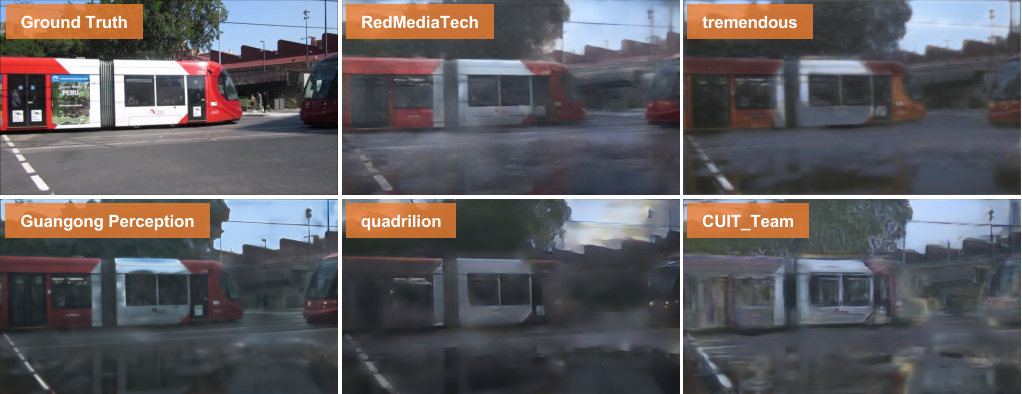}
  \caption{Visual comparison on a rainy scene. From left to right, top to bottom: Ground Truth, RedMediaTech, tremendous, Guangong Perception, quadrillion, and CUIT\_Team.}
  \label{fig:vis_rainy}
\end{figure*}

\begin{figure*}[t]
  \centering
  \includegraphics[width=\linewidth]{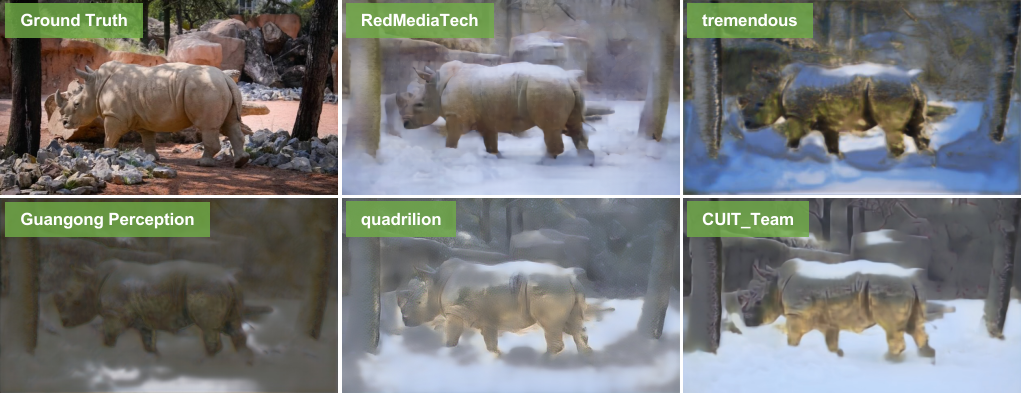}
  \caption{Visual comparison on a snowy scene. From left to right, top to bottom: Ground Truth, RedMediaTech, tremendous, Guangong Perception, quadrillion, and CUIT\_Team.}
  \label{fig:vis_snowy}
\end{figure*}

Visual comparisons of the restored results from all five teams on a rainy scene and a snowy scene are shown in \cref{fig:vis_rainy} and \cref{fig:vis_snowy}, respectively.

\section{Teams and Methods}
\label{sec:methods}

This section describes the methods proposed by the five participating teams, ordered by their final ranking.

\subsection{RedMediaTech}
\label{sec:redmediatech}

\begin{figure}[t]
  \centering
  \includegraphics[width=\linewidth]{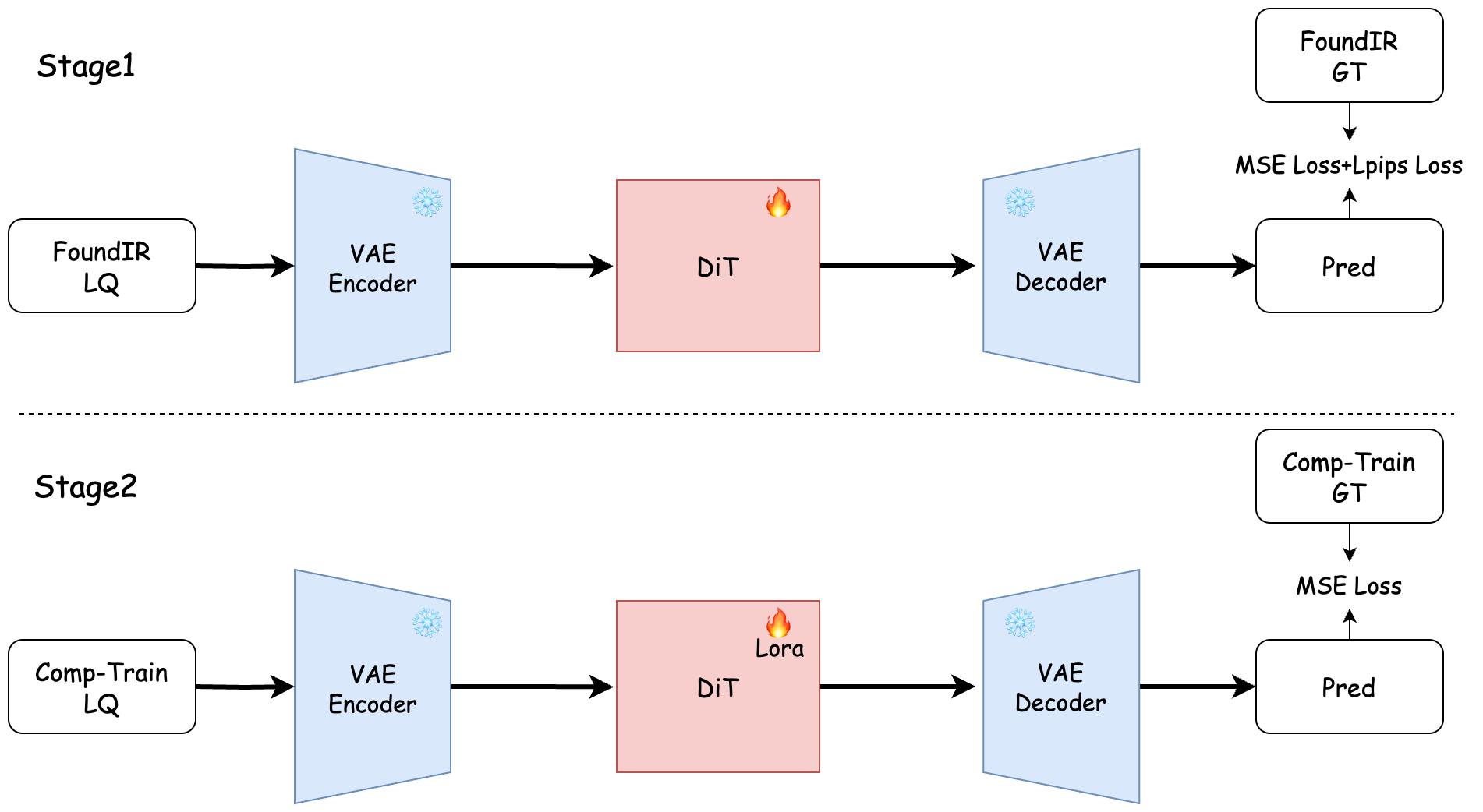}
  \caption{Overview of the RedMediaTech method. Stage 1: the DiT backbone is pre-trained on FoundIR with MSE and LPIPS losses. Stage 2: the DiT is fine-tuned on the competition data using LoRA with MSE loss. The VAE remains frozen throughout.}
  \label{fig:redmediatech}
\end{figure}

This team builds upon the pre-trained FLUX.2 \cite{flux2} [klein] 4B architecture, designing a diffusion-transformer-based image restoration framework operating in latent space. The VAE is frozen throughout training and inference, while the central DiT module is optimized to learn degradation-aware restoration priors. This design preserves the stable latent encoding/decoding capability of the pre-trained backbone while allowing the transformer to adapt to the target restoration task. Notably, this is a single-frame model that does not utilize temporal information. The overall pipeline is illustrated in \cref{fig:redmediatech}.

\textbf{Network design.}
The overall network consists of a frozen VAE and a trainable DiT backbone. The input degraded image is first encoded into latent space by the VAE encoder, after which the latent features are processed by the DiT for restoration-oriented denoising and refinement. The restored latent representation is then decoded by the frozen VAE decoder to obtain the final output image. In the first stage, full training of the DiT parameters is performed on external data. In the second stage, a parameter-efficient LoRA-based fine-tuning strategy is adopted on the DiT, reducing computational overhead while improving task adaptation.

\textbf{Training details.}
The training procedure consists of two stages:
\begin{itemize}
    \item \textbf{Stage 1: Large-scale pre-training.} The model is initialized from the pre-trained FLUX.2 [klein] 4B checkpoint. The VAE is frozen and the DiT is fully trained with a learning rate of $1 \times 10^{-4}$ and batch size of 16. The loss function combines MSE and LPIPS losses. This stage is trained on the FoundIR dataset~\cite{foundir} for 110,000 iterations.
    \item \textbf{Stage 2: Competition-specific fine-tuning.} The model is fine-tuned on the official competition training set using LoRA-based adaptation on the DiT. The batch size is 8, the loss function is standard MSE, and the stage is trained for 25,000 iterations.
\end{itemize}

\textbf{Testing details.}
During inference, the degraded input image is encoded into latent space by the frozen VAE, processed by the fine-tuned DiT, and decoded back to image space. No additional external models are required at test time.

\subsection{tremendous}
\label{sec:tremendous}

\begin{figure}[t]
  \centering
  \includegraphics[width=\linewidth]{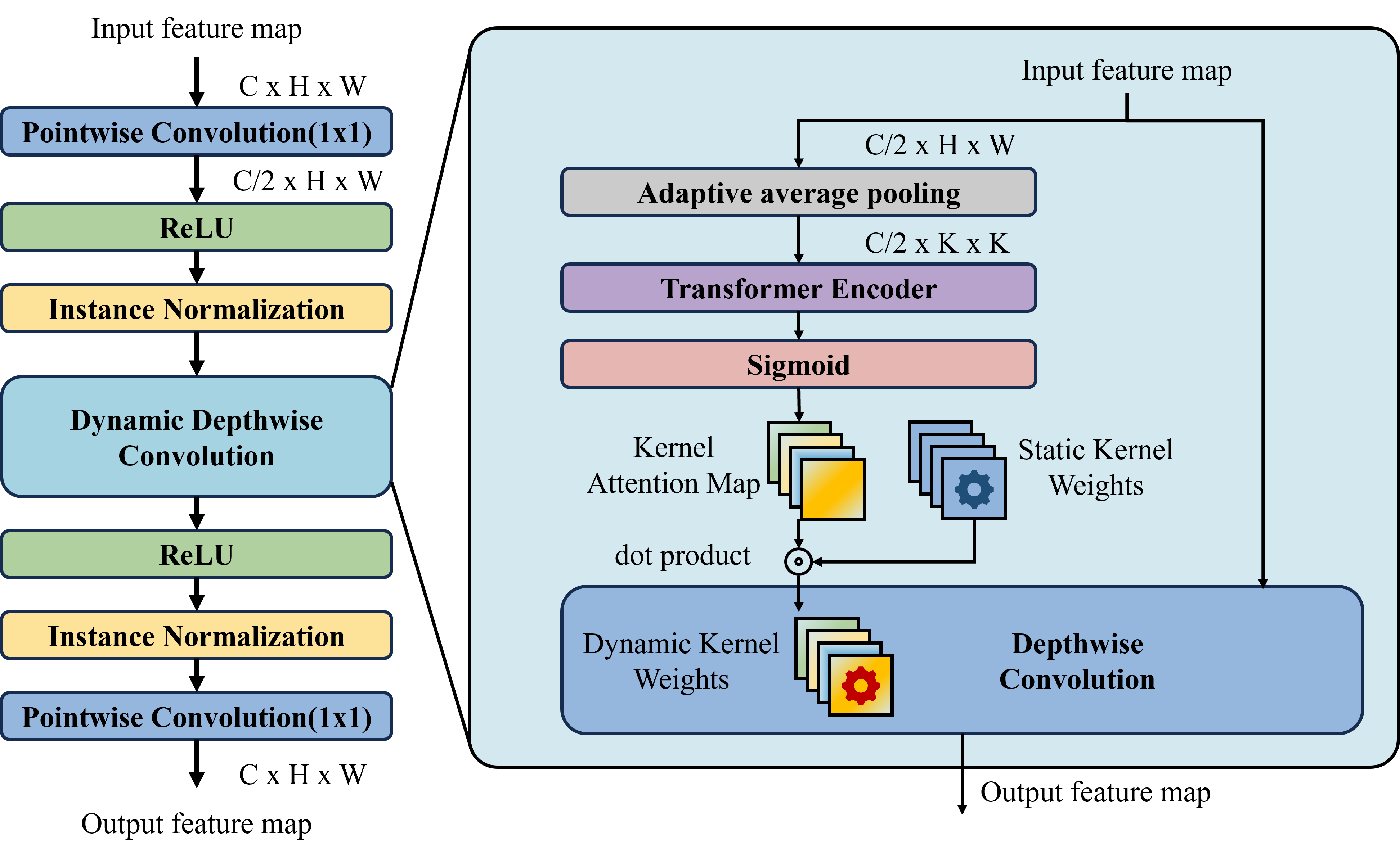}
  \caption{Architecture of the TDyConv module proposed by team tremendous. A Transformer encoder generates kernel attention maps that dynamically modulate depthwise convolution weights, enabling content-adaptive processing of heterogeneous weather degradations.}
  \label{fig:tremendous}
\end{figure}

This team proposes DyStd-Net, which employs dynamic convolutions guided by a Transformer module (TDyConv) to construct a content-adaptive video frame restoration paradigm. The approach targets the significant heterogeneity among degradation patterns such as rain streaks, snowflakes, and haze by establishing a collaborative modeling paradigm of ``global semantic-guided local feature extraction.'' The architecture of the TDyConv module is shown in \cref{fig:tremendous}.

\textbf{Network design.}
The overall network follows a U-Net-like~\cite{ronneberger2015u} multi-scale encoder-decoder structure. Through hierarchical residual connections and downsampling modules, a deep feature pyramid is constructed, capable of capturing multi-level information ranging from fine rain/snow texture details to background semantic structures. The core innovation is the TDyConv module, which utilizes a Transformer~\cite{vaswani2017attention} to capture long-range contextual dependencies and generate refined kernel attention maps. This mechanism enables dynamic modulation of convolution weights based on input content, allowing the network to flexibly handle heterogeneous degradation modes. To mitigate checkerboard artifacts, convolutional layers with a stride of 2 are employed for downsampling, paired with bilinear interpolation for upsampling. The network comprises 6.797M parameters with a computational cost of 69.706 GFLOPs.

\textbf{Training details.}
The Adam optimizer ($\beta_1 = 0.95$, $\beta_2 = 0.999$) is employed with weight decay. The initial learning rate is $1 \times 10^{-4}$, with a total of 150 epochs and batch size of 4. A progressive training strategy with staged resolutions is implemented, where the resolution is gradually increased from $256 \times 256$ to $832 \times 480$. The joint loss function is:
\begin{equation}
    L_{\text{total}} = L_1 + 0.1 \times L_{\text{SSIM}} + 0.1 \times L_{\text{Perceptual}},
\end{equation}
where $L_1$ is Smooth L1 loss, $L_{\text{SSIM}}$ is SSIM loss, and $L_{\text{Perceptual}}$ is VGG16-based~\cite{simonyan2014very} perceptual loss. Data augmentation includes random cropping (scales $0.5 \sim 1.5$), flipping, and multi-angle rotation. All experiments are conducted on an NVIDIA RTX 2080Ti GPU.

\textbf{Testing details.}
Inference is performed at the full resolution of $832 \times 480$.

\subsection{Guangong Perception}
\label{sec:gongguan}

\begin{figure*}[t]
  \centering
  \includegraphics[width=\linewidth]{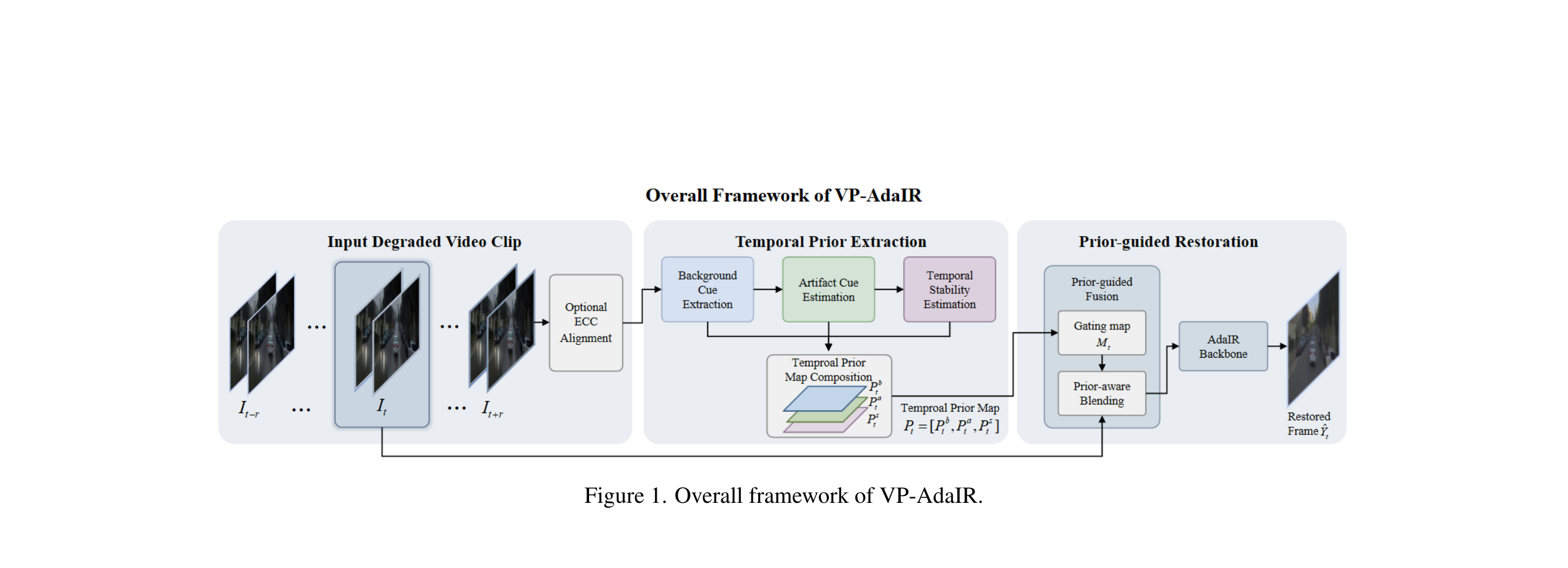}
  \caption{Overall framework of VP-AdaIR proposed by team Guangong Perception. The method takes a degraded video clip as input, extracts a three-channel temporal prior map (background, artifact, and temporal stability cues), and injects it into the AdaIR backbone through prior-guided fusion.}
  \label{fig:gongguan}
\end{figure*}

This team proposes VP-AdaIR, which extends the AdaIR~\cite{cui2025adair} image restoration backbone to video-frame restoration for adverse weather degradation, especially snowy scenes. The key idea is to design a three-channel temporal prior map from degraded video clips and inject it into the AdaIR backbone as prior guidance. The overall framework is illustrated in \cref{fig:gongguan}.

\textbf{Network design.}
Given a degraded video clip, the center frame is taken as the restoration target and its neighboring frames serve as auxiliary temporal context. An optional ECC alignment step is applied to reduce the influence of camera motion. A temporal prior extraction stage then estimates three complementary cues:
\begin{itemize}
    \item \textbf{Background cue:} extracts stable background structures from neighboring frames.
    \item \textbf{Artifact cue:} identifies weather-related degradation patterns.
    \item \textbf{Temporal stability cue:} estimates inter-frame consistency.
\end{itemize}
These three cues are composed into a three-channel temporal prior map, which is fed into a prior-guided fusion module and injected into the AdaIR backbone through a gating mechanism and prior-aware blending to produce the restored frame.

\textbf{Training details.}
The model is trained using a supervised training scheme with degraded video clips and their corresponding clean target frames. The temporal prior map is generated from neighboring frames and fused with the degraded center frame inside the network. In addition to the competition-provided data, 20 snowy scenes from the VideoDesnowing dataset~\cite{chen2023snow} are used as additional training data. These extra data are used to construct degraded-clean frame pairs for training and fine-tuning. Neighboring frames are organized into short temporal clips, and optional alignment is performed to improve the quality of temporal prior extraction.

\textbf{Testing details.}
During inference, the method takes a degraded video clip as input, extracts the three-channel temporal prior map from neighboring frames, injects the prior into the AdaIR backbone through the prior-guided fusion module, and outputs the restored target frame. Runtime is approximately 0.54 seconds per image on GPU.

\subsection{quadrillion}
\label{sec:quadrillion}

\begin{figure*}[t]
  \centering
  \includegraphics[width=\linewidth]{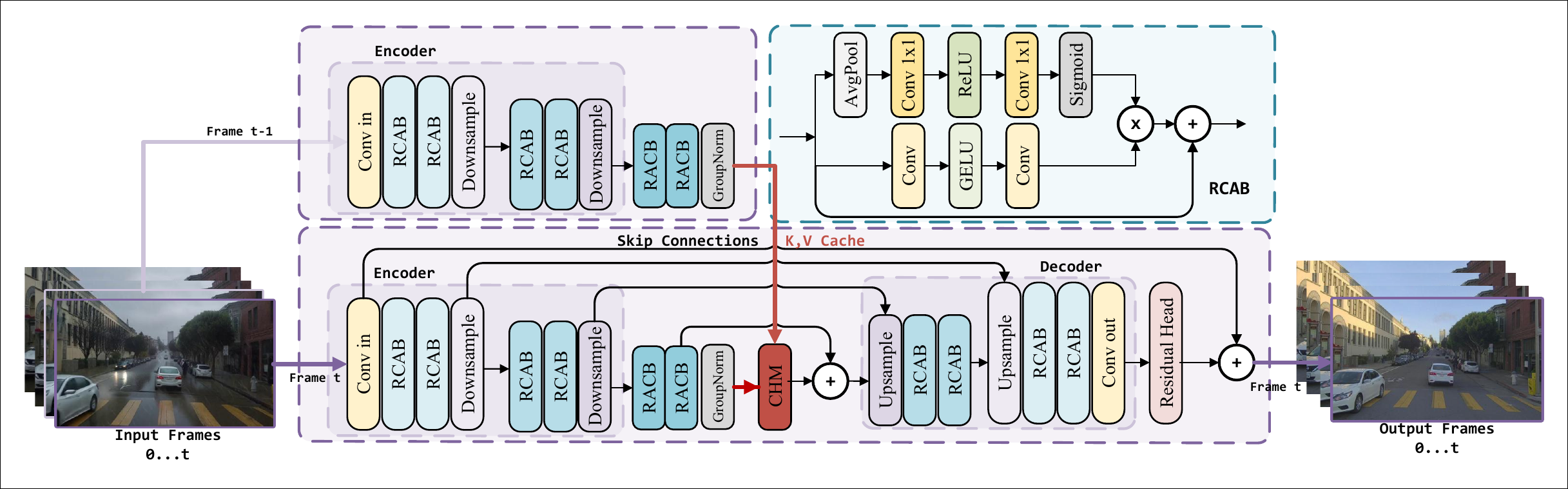}
  \caption{Overall architecture of EF$^3$Net proposed by team quadrillion. A Siamese Historyless Encoder independently processes previous and current frames. A Causal History Model (CHM) dynamically fuses temporal memory via Key/Value caching at the bottleneck.}
  \label{fig:quadrillion}
\end{figure*}

This team proposes EF$^3$Net (Efficient-fusion Focal Frequency Network), a lightweight yet robust video restoration framework that leverages frequency-aware and perceptual optimization rather than relying solely on rigid pixel-level supervision. The method addresses the severe spatial misalignment between degraded inputs and clean ground-truth targets in the composite weather dataset. The overall architecture is shown in \cref{fig:quadrillion}.

\textbf{Network design.}
The architecture consists of a Siamese Historyless Encoder combined with Late Fusion. Given a video sequence, the previous frame $I_{t-1}$ and the current degraded frame $I_t$ are processed independently by a shared-weight encoder. The encoder follows a UNet-like structure integrated with Residual Channel Attention Blocks (RCAB)~\cite{zhang2018image}. At the deepest bottleneck, a Causal History Model (CHM)~\cite{ghasemabadi2024learning} dynamically fuses temporal memory via Key/Value caching. The CHM extracts deep features of $I_{t-1}$ to formulate a temporal memory that dynamically guides and fuses with the high-level features of $I_t$. The decoder uses bilinear upsampling followed by $3 \times 3$ convolutions to prevent checkerboard artifacts. A dedicated residual head predicts the degradation residual, scaled by a constant factor $\alpha = 0.3$.

\textbf{Training details.}
The EF$^3$Net is trained end-to-end on a single NVIDIA GeForce RTX 3090 GPU (24GB VRAM). The training pipeline is built upon the BasicSR~\cite{basicsr} framework and adapted from the TURTLE~\cite{ghasemabadi2024learning} codebase. The AdamW optimizer is employed with a weight decay of $1 \times 10^{-4}$, $\beta_1 = 0.9$, $\beta_2 = 0.99$, and $\epsilon = 1 \times 10^{-6}$. The learning rate starts at $4 \times 10^{-4}$ and decays to $1 \times 10^{-7}$ following a Cosine Annealing schedule over 30,000 iterations. Spatial patches of size $256 \times 256$ are extracted with a batch size of 4. Standard data augmentations including random flips and rotations are applied. Automatic Mixed Precision (AMP) is used and gradient norms are clipped to 1.0.

The loss function is a weighted combination:
\begin{multline}
    \mathcal{L}_{\text{total}} = \lambda_1 \mathcal{L}_{L1} + \lambda_{\text{vgg}} \mathcal{L}_{\text{VGG}} \\
    + \lambda_{\text{fft}} \mathcal{L}_{\text{FFT}} + \lambda_{\text{ffl}} \mathcal{L}_{\text{FFL}},
\end{multline}
where the weights are set to $\lambda_1 = 0.3$, $\lambda_{\text{vgg}} = 1.0$, $\lambda_{\text{fft}} = 0.1$, and $\lambda_{\text{ffl}} = 800.0$. The Focal Frequency Loss ($\mathcal{L}_{\text{FFL}}$)~\cite{jiang2021focal} is weighted exceptionally high to aggressively enforce structural alignment in the frequency domain.
\begin{figure*}[t]
  \centering
  \includegraphics[width=\linewidth]{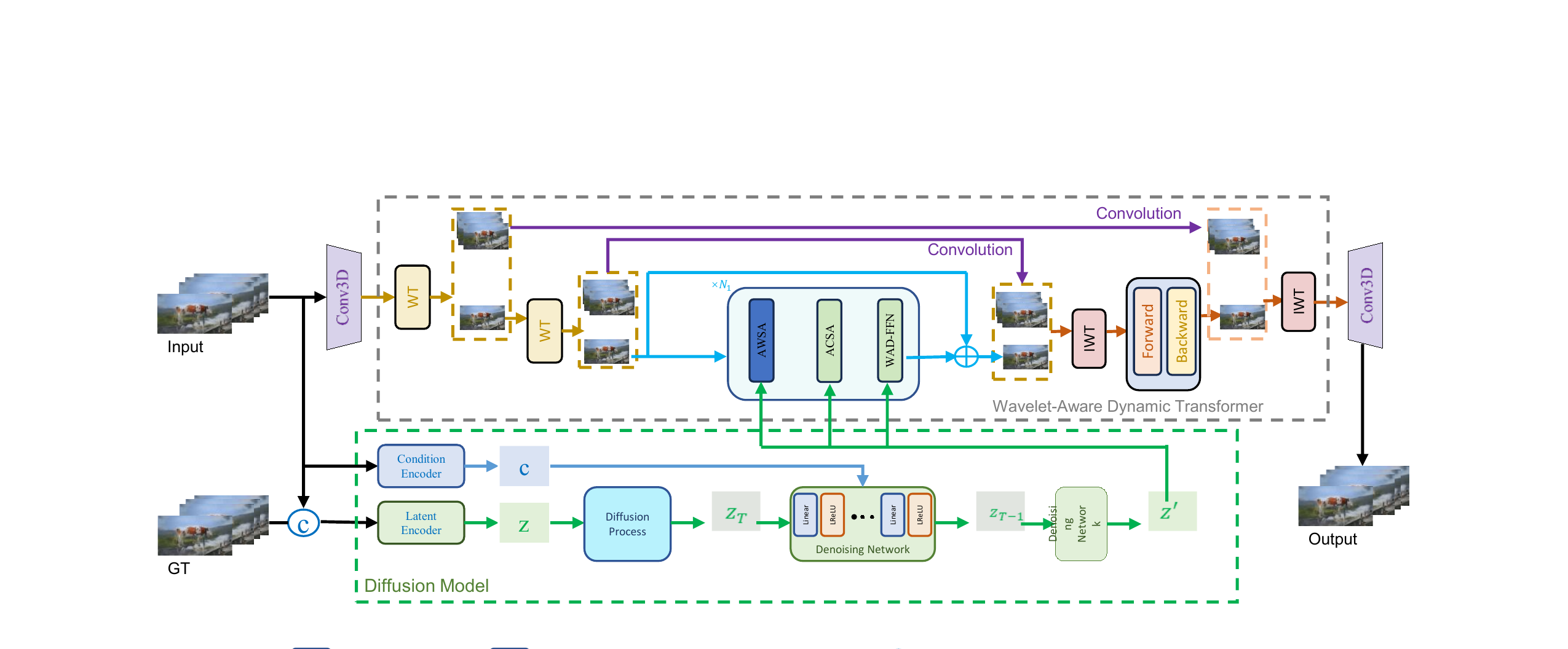}
  \caption{Overall architecture of VD-Diff proposed by CUIT\_Team. The framework consists of a conditional diffusion model for implicit prior generation and a wavelet-decomposition-based restoration backbone with Adaptive Window Self-Attention (AWSA), Adaptive Channel Self-Attention (ACSA), and bidirectional temporal propagation.}
  \label{fig:cuit}
\end{figure*}

\textbf{Testing details.}
Video sequences are processed chronologically. For a given time step $t$, the current frame $I_t$ and the previous frame $I_{t-1}$ are fed independently into the Siamese encoders. The temporal memory is passed dynamically via the KV cache of the CHM. The network predicts the degradation residual, and the final restored output is clamped to $[0, 1]$. The computational complexity is 127.48 GFLOPs with 7.46M parameters, achieving approximately 0.05 seconds per frame on an NVIDIA RTX 4070 GPU with a peak memory of 0.20 GB.

\subsection{CUIT\_Team}
\label{sec:cuit}

This team proposes VD-Diff, a three-stage diffusion-prior-guided restoration framework for adverse-weather video restoration. The method follows the VD-Diff~\cite{rao2025rethinking} framework, with the main contribution being redesigned attention mechanisms in the restoration backbone. The overall architecture is shown in \cref{fig:cuit}.

\textbf{Network design.}
The overall framework consists of three components: a teacher prior extractor, a conditional diffusion model, and a prior-guided video restoration backbone. The teacher prior extractor learns a compact implicit restoration prior from degraded frames and their corresponding ground truth. The conditional diffusion model learns to predict this implicit prior directly from degraded inputs during inference, when ground truth is unavailable. The restoration backbone adopts a wavelet decomposition + adaptive Transformer + bidirectional temporal propagation architecture. The input video is first processed by a 3D convolution layer to extract shallow spatio-temporal features, then decomposed into low-frequency structural components and high-frequency detail components through two-level Haar wavelet decomposition. The core component is the Wavelet-Aware Dynamic Transformer Layer (WADTL), where each block consists of Adaptive Window Self-Attention (AWSA), Adaptive Channel Self-Attention (ACSA), and a prior-modulated Feed-Forward network. The implicit prior $k_v$ is introduced into all three submodules for conditional modulation.

\textbf{Training details.}
The training consists of three stages with learning rates of $1 \times 10^{-4}$, $5 \times 10^{-5}$, and $1 \times 10^{-5}$, respectively. The AdamW optimizer is used with a batch size of 2. Each training sample contains 10 consecutive video frames. In addition to the officially provided dataset, 1,430 images from the dataset of Chen~\etal~\cite{chen2018robust} are used as extra training data.
\begin{itemize}
    \item \textbf{Stage I: Teacher Prior Learning.} The teacher prior extractor and the restoration backbone are trained jointly so that the restoration network first establishes a stable mapping under ideal prior guidance.
    \item \textbf{Stage II: Diffusion Prior Fitting.} The conditional diffusion model is trained to approximate the teacher prior space learned in Stage I.
    \item \textbf{Stage III: Joint Optimization.} Both the diffusion branch and the restoration branch are jointly optimized, progressively improving compatibility between the diffusion prior and the restoration backbone.
\end{itemize}

\textbf{Testing details.}
During inference, only the pipeline of diffusion-prior generation followed by restoration-backbone reconstruction is retained. The prior is directly predicted by the diffusion model from degraded inputs and fed into the restoration backbone, enabling video restoration without requiring ground truth. The runtime is approximately 0.13 seconds per image.

\section*{Acknowledgments}
This work was partially supported by the Postdoctoral Fellowship Program of CPSF under Grant Number GZC20252293, the China Postdoctoral Science Foundation-Anhui Joint Support Program under Grant Number 2024T017AH, China Postdoctoral Science Foundation under Grant Number  2025M783529, Anhui Postdoctoral Scientific Research Program Foundation (No.2025A1015), the Fundamental Research Funds for the Central Universities (No. WK2100250064). We thank the LoViF 2026 sponsors: Tencent and IMCL lab. We thank all participants for their contributions and the CodaBench platform for hosting the competition.

\appendix

\section{Teams and Affiliations}
\label{sec:affiliations}

\subsection*{LoViF 2026 Challenge Organizers}

\noindent\textit{Title:} LoViF 2026 Challenge on Weather Removal in Videos\\
\textit{Members:} Chenghao Qian (\href{mailto:yohji.qian@gmail.com}{yohji.qian@gmail.com}), Xin Li(\href{mailto:xin.li@ustc.edu.cn}{xin.li@ustc.edu.cn}), Yeying Jin(\href{mailto:jinyeying@u.nus.edu}{jinyeying@u.nus.edu}), Shangquan Sun (\href{mailto:shangquan.sun@ntu.edu.sg}{shangquan.sun@ntu.edu.sg})\\
\textit{Affiliations:} University of Leeds; University of Science and Technology of China; Tencent; Nanyang Technological University

\subsection*{RedMediaTech}

\noindent\textit{Title:} Latent Diffusion Transformer for Weather Removal\\
\textit{Members:} Yilian Zhong (\href{mailto:zhongyilian@fudan.edu.cn}{zhongyilian@fudan.edu.cn}), Yuxiang Chen, Shibo Yin, Yushun Fang, Xilei Zhu, Yahui Wang, Chen Lu\\
\textit{Affiliations:} Xiaohongshu Inc

\subsection*{tremendous}

\noindent\textit{Title:} DyStd-Net: Dynamic Convolution-based Video Weather Removal\\
\textit{Members:} Zhuohang Shi (\href{mailto:shi_zhuo_hang@yeah.net}{shi\_zhuo\_hang@yeah.net}), Xiaojing Liu, Jiao Liu, Yatong Zhou, Shuai Liu\\
\textit{Affiliations:} Hebei University of Technology

\subsection*{Guangong Perception}

\noindent\textit{Title:} VP-AdaIR: Temporal Prior-Guided Video Desnowing and Weather Removal\\
\textit{Members:} Qiang Deng (\href{mailto:derricksz@163.com}{derricksz@163.com}), Jiajia Mi, Qianhao Luo, Weiling Li\\
\textit{Affiliations:} Dongguan University of Technology; Guangdong University of Technology

\subsection*{quadrillion}

\noindent\textit{Title:} EF$^3$Net: Frequency-Aware Network for Video Weather Removal\\
\textit{Members:} Junyang Jiang (\href{mailto:jiangjunyang@mails.gdut.edu.cn}{jiangjunyang@mails.gdut.edu.cn}), Yuping Sun\\
\textit{Affiliations:} Guangdong University of Technology

\subsection*{CUIT\_Team}

\noindent\textit{Title:} VD-Diff: Diffusion-Prior-Guided Video Weather Restoration\\
\textit{Members:} Ying Fu (\href{mailto:fuying@cuit.edu.cn}{fuying@cuit.edu.cn}), Jianan Tian, Jifan Zhang, Chen Zhou\\
\textit{Affiliations:} Chengdu University of Information Technology

{
    \small
    \bibliographystyle{ieeenat_fullname}
    \bibliography{main}

@String(CVPR  = {CVPR})

@String(ICCV  = {ICCV})

@String(ECCV  = {ECCV})

@String(NIPS  = {NeurIPS})

@String(ICLR  = {ICLR})

@String(TIP   = {IEEE TIP})

@String(CVPRW = {CVPRW})

@inproceedings{lovif2026realir,
  title={{LoViF} 2026 Challenge on Real-World All-in-One Image Restoration: Methods and Results},
  author={Chen, Xiang and Li, Hao and Dong, Jiangxin and Pan, Jinshan and Li, Xin and others},
  booktitle={Proceedings of the IEEE/CVF Conference on Computer Vision and Pattern Recognition (CVPR) Workshops},
  year={2026}
}

@inproceedings{lovif2026MQualityScoring,
  title={The 1st {LoViF} Challenge on Efficient VLM for Multimodal Creative Quality Scoring: Methods and Results},
  author={Zhang, Jusheng and Lyu, Qinhan and Ma,  Sizhuo and Cao, Sheng and Wang, Jian and Li, Xin and Wang, Keze and Zheng, Yongsen and Yang, Jing and others},
  booktitle={Proceedings of the IEEE/CVF Conference on Computer Vision and Pattern Recognition (CVPR) Workshops},
  year={2026}
}

@inproceedings{lovif2026WeatherRemoval,
  title={{LoViF} 2026 The First Challenge on Weather Removal in Videos},
  author={Qian, Chenghao and Li, Xin and Jin, Yeying and Sun, Shangquan and others},
  booktitle={Proceedings of the IEEE/CVF Conference on Computer Vision and Pattern Recognition (CVPR) Workshops},
  year={2026}
}

@inproceedings{lovif2026HQA,
  title={{LoViF} 2026 The First Challenge on Holistic Quality Assessment for 4D World Model (PhyScore)},
  author={Luo, Wei and Lu, Yiting and  Li, Xin and Li, Haoran and Guan, Fengbin and Gao, Chen and Jin, Xin and Li, Yong and Chen, Zhibo	and others},
  booktitle={Proceedings of the IEEE/CVF Conference on Computer Vision and Pattern Recognition (CVPR) Workshops},
  year={2026}
}

@inproceedings{lovif2026SeIQA,
  title={{LoViF} 2026 The First Challenge on Human-Oriented Semantic Image Quality Assessment: Methods and Results},
  author={Li, Xin and Xu, Daoli	and Luo,  Wei and Xiang, Guoqiang and Li, Haoran and Zhuang, Chengyu and Chen, Zhibo and Guan, Jian and Li, Weipingand others},
  booktitle={Proceedings of the IEEE/CVF Conference on Computer Vision and Pattern Recognition (CVPR) Workshops},
  year={2026}
}

@inproceedings{li2025ntire,
  title={NTIRE 2025 challenge on day and night raindrop removal for dual-focused images: Methods and results},
  author={Li, Xin and Jin, Yeying and Jin, Xin and Wu, Zongwei and Li, Bingchen and Wang, Yufei and Yang, Wenhan and Li, Yu and Chen, Zhibo and Wen, Bihan and others},
  booktitle={Proceedings of the Computer Vision and Pattern Recognition Conference},
  pages={1172--1183},
  year={2025}
}

@inproceedings{ntire26dual_focus, 
title={{    NTIRE 2026 The Second Challenge on Day and Night Raindrop Removal for Dual-Focused Images: Methods and Results    }}, 
author={    Li, Xin and  Jin, Yeying and  Yao, Suhang and  Lin, Beibei and  Fan, Zhaoxin and   Yan, Wending and  Jin, Xin and  Wu, Zongwei  and  Li, Bingchen  and  Shi, Peishu and  Yang, Yufei and  Li, Yu and  Chen, Zhibo  and  Wen, Bihan and  Tan, Robby and  Timofte, Radu and others    },   
booktitle={Proceedings of the IEEE/CVF Conference on Computer Vision and Pattern Recognition (CVPR) Workshops},  
year = {2026} 
}

@inproceedings{jin2024raindrop_clarity,
  title={Raindrop Clarity: A Dual-Focused Dataset for Day and Night Raindrop Removal},
  author={Jin, Yeying and Li, Xin and Wang, Jiadong and Zhang, Yan and Zhang, Malu},
  booktitle={ECCV},
  pages={1--17},
  year={2024},
  organization={Springer}
}

@article{qian3d,
  title={3D Weather Editing with 4D Gaussian Field},
  author={Qian, Chenghao and Li, Wenjing and Guo, Yuhu and Markkula, Gustav}
}

@inproceedings{barnes2020oxford,
  title={The oxford radar robotcar dataset: A radar extension to the oxford robotcar dataset},
  author={Barnes, Dan and Gadd, Matthew and Murcutt, Paul and Newman, Paul and Posner, Ingmar},
  booktitle={2020 IEEE international conference on robotics and automation (ICRA)},
  pages={6433--6438},
  year={2020},
  organization={IEEE}
}

@inproceedings{varanka2026zero,
  title={Zero-Shot Video Deraining with Video Diffusion Models},
  author={Varanka, Tuomas and Gonzalez, Juan Luis and Kim, Hyeongwoo and Garrido, Pablo and Yao, Xu},
  booktitle={Proceedings of the IEEE/CVF Winter Conference on Applications of Computer Vision},
  pages={677--687},
  year={2026}
}

@inproceedings{qian2024allweather,
  title={Allweather-net: Unified image enhancement for autonomous driving under adverse weather and low-light conditions},
  author={Qian, Chenghao and Rezaei, Mahdi and Anwar, Saeed and Li, Wenjing and Hussain, Tanveer and Azarmi, Mohsen and Wang, Wei},
  booktitle={International Conference on Pattern Recognition},
  pages={151--166},
  year={2024},
  organization={Springer}
}

@article{zhang2025egvd,
  title={Egvd: Event-guided video deraining},
  author={Zhang, Yueyi and Wang, Jin and Weng, Wenming and Sun, Xiaoyan and Xiong, Zhiwei},
  journal={IEEE Transactions on Neural Networks and Learning Systems},
  year={2025},
  publisher={IEEE}
}

@article{qian2025weatherdg,
  title={WeatherDG: LLM-assisted procedural weather generation for domain-generalized semantic segmentation},
  author={Qian, Chenghao and Guo, Yuhu and Mo, Yuhong and Li, Wenjing},
  journal={IEEE Robotics and Automation Letters},
  year={2025},
  publisher={IEEE}
}

@inproceedings{qian2026weatheredit,
  title={Weatheredit: Controllable weather editing with 4d gaussian field},
  author={Qian, Chenghao and Li, Wenjing and Guo, Yuhu and Markkula, Gustav},
  booktitle={Proceedings of the AAAI Conference on Artificial Intelligence},
  volume={40},
  number={10},
  pages={8511--8519},
  year={2026}
}

@inproceedings{qian2025weathergs,
  title={Weathergs: 3d scene reconstruction in adverse weather conditions via gaussian splatting},
  author={Qian, Chenghao and Guo, Yuhu and Li, Wenjing and Markkula, Gustav},
  booktitle={2025 IEEE International Conference on Robotics and Automation (ICRA)},
  pages={185--191},
  year={2025},
  organization={IEEE}
}

@inproceedings{sakaridis2018semantic,
  title={Semantic foggy scene understanding with synthetic data},
  author={Sakaridis, Christos and Dai, Dengxin and Van Gool, Luc},
  booktitle={International Journal of Computer Vision},
  volume={126},
  pages={973--992},
  year={2018},
  publisher={Springer}
}

@inproceedings{jiang2020multi,
  title={Multi-scale progressive fusion network for single image deraining},
  author={Jiang, Kui and Wang, Zhongyuan and Yi, Peng and Chen, Chen and Huang, Baojin and Luo, Yimin and Ma, Jiayi and Jiang, Junjun},
  booktitle=CVPR,
  pages={8346--8355},
  year={2020}
}

@inproceedings{zamir2021multi,
  title={Multi-stage progressive image restoration},
  author={Zamir, Syed Waqas and Arora, Aditya and Khan, Salman and Hayat, Munawar and Khan, Fahad Shahbaz and Yang, Ming-Hsuan and Shao, Ling},
  booktitle=CVPR,
  pages={14821--14831},
  year={2021}
}

@inproceedings{liu2018desnownet,
  title={DesnowNet: Context-aware deep network for snow removal},
  author={Liu, Yun-Fu and Jaw, Da-Wei and Huang, Shih-Chia and Hwang, Jenq-Neng},
  booktitle=TIP,
  volume={27},
  number={6},
  pages={3064--3073},
  year={2018}
}

@inproceedings{chen2021all,
  title={All snow removed: Single image desnowing algorithm using hierarchical dual-tree complex wavelet representation and contradict channel loss},
  author={Chen, Wei-Ting and Fang, Hao-Yu and Hsieh, Cheng-Lin and Tsai, Cheng-Che and Chen, I and Ding, Jian-Jiun and Kuo, Sy-Yen},
  booktitle=ICCV,
  pages={4196--4205},
  year={2021}
}

@inproceedings{qin2020ffa,
  title={FFA-Net: Feature fusion attention network for single image dehazing},
  author={Qin, Xu and Wang, Zhilin and Bai, Yuanchao and Xie, Xiaodong and Jia, Huizhu},
  booktitle={AAAI},
  pages={11908--11915},
  year={2020}
}

@inproceedings{song2023vision,
  title={Vision transformers for single image dehazing},
  author={Song, Yuda and He, Zhuqing and Qian, Hui and Du, Xin},
  booktitle=TIP,
  volume={32},
  pages={1927--1941},
  year={2023}
}

@inproceedings{xue2019video,
  title={Video enhancement with task-oriented flow},
  author={Xue, Tianfan and Chen, Baian and Wu, Jiajun and Wei, Donglai and Freeman, William T},
  booktitle={International Journal of Computer Vision},
  volume={127},
  pages={1106--1125},
  year={2019},
  publisher={Springer}
}

@inproceedings{wang2019edvr,
  title={EDVR: Video restoration with enhanced deformable convolutional networks},
  author={Wang, Xintao and Chan, Kelvin CK and Yu, Ke and Dong, Chao and Change Loy, Chen},
  booktitle=CVPRW,
  pages={1954--1963},
  year={2019}
}

@inproceedings{chan2022basicvsr++,
  title={BasicVSR++: Improving video super-resolution with enhanced propagation and alignment},
  author={Chan, Kelvin CK and Zhou, Shangchen and Xu, Xiangyu and Loy, Chen Change},
  booktitle=CVPR,
  pages={5972--5981},
  year={2022}
}

@inproceedings{yang2020self,
  title={Self-learning video rain streak removal: When cyclic consistency meets temporal correspondence},
  author={Yang, Wenhan and Tan, Robby T and Wang, Shiqi and Fang, Yuming and Liu, Jiaying},
  booktitle=CVPR,
  pages={1720--1729},
  year={2020}
}

@inproceedings{chen2023snow,
  title={Snow removal in video: A new dataset and a novel method},
  author={Chen, Haoyu and Ren, Jingjing and Gu, Jinjin and Wu, Hongtao and Lu, Xuequan and Cai, Haoming and Zhu, Lei},
  booktitle=ICCV,
  pages={13165--13176},
  year={2023}
}

@inproceedings{yang2023video,
  title={Video adverse-weather-component suppression network via weather messenger and adversarial backpropagation},
  author={Yang, Yijun and Yang, Wenhan and Tan, Robby T},
  booktitle=ICCV,
  pages={13200--13210},
  year={2023}
}

@article{chih2025weatherweaver,
  title={Controllable weather synthesis and removal with video diffusion models},
  author={Chih-Hao, Lin and others},
  journal={arXiv preprint arXiv:2505.00704},
  year={2025}
}

@misc{lovif2026,
  author={{LoViF Organizing Committee}},
  title={The 1st Workshop on Low-level Vision Frontiers (LoViF) at CVPR 2026},
  howpublished={\url{https://lovif-cvpr2026-workshop.github.io/}},
  year={2026}
}

@article{wang2004image,
  title={Image quality assessment: From error visibility to structural similarity},
  author={Wang, Zhou and Bovik, Alan C and Sheikh, Hamid R and Simoncelli, Eero P},
  journal=TIP,
  volume={13},
  number={4},
  pages={600--612},
  year={2004}
}

@inproceedings{zhang2018unreasonable,
  title={The unreasonable effectiveness of deep features as a perceptual metric},
  author={Zhang, Richard and Isola, Phillip and Efros, Alexei A and Shechtman, Eli and Wang, Oliver},
  booktitle=CVPR,
  pages={586--595},
  year={2018}
}

@misc{flux2,
  title={FLUX.2: Frontier Visual Intelligence},
  author={{Black Forest Labs}},
  howpublished={\url{https://bfl.ai/blog/flux-2}},
  year={2025}
}

@inproceedings{foundir,
  title={FoundIR: Unleashing million-scale training data to advance foundation models for image restoration},
  author={Li, Hao and Chen, Xiang and Dong, Jiangxin and Tang, Jinhui and Pan, Jinshan},
  booktitle=ICCV,
  pages={12626--12636},
  year={2025}
}

@inproceedings{ronneberger2015u,
  title={U-Net: Convolutional networks for biomedical image segmentation},
  author={Ronneberger, Olaf and Fischer, Philipp and Brox, Thomas},
  booktitle={MICCAI},
  pages={234--241},
  year={2015},
  publisher={Springer}
}

@inproceedings{vaswani2017attention,
  title={Attention is all you need},
  author={Vaswani, Ashish and Shazeer, Noam and Parmar, Niki and Uszkoreit, Jakob and Jones, Llion and Gomez, Aidan N and Kaiser, {\L}ukasz and Polosukhin, Illia},
  booktitle=NIPS,
  volume={30},
  year={2017}
}

@article{simonyan2014very,
  title={Very deep convolutional networks for large-scale image recognition},
  author={Simonyan, Karen and Zisserman, Andrew},
  journal={arXiv preprint arXiv:1409.1556},
  year={2014}
}

@inproceedings{cui2025adair,
  title={AdaIR: Adaptive all-in-one image restoration via frequency mining and modulation},
  author={Cui, Yuning and Zamir, Syed Waqas and Khan, Salman and Knoll, Alois and Shah, Mubarak and Khan, Fahad Shahbaz},
  booktitle=ICLR,
  pages={57335--57356},
  year={2025}
}

@inproceedings{zhang2018image,
  title={Image super-resolution using very deep residual channel attention networks},
  author={Zhang, Yulun and Li, Kunpeng and Li, Kai and Wang, Lichen and Zhong, Bineng and Fu, Yun},
  booktitle=ECCV,
  pages={294--310},
  year={2018}
}

@inproceedings{ghasemabadi2024learning,
  title={Learning truncated causal history model for video restoration},
  author={Ghasemabadi, Amirhosein and Janjua, Muhammad Kamran and Salameh, Mohammad and Niu, Di},
  booktitle=NIPS,
  year={2024}
}

@inproceedings{jiang2021focal,
  title={Focal frequency loss for image reconstruction and synthesis},
  author={Jiang, Liming and Dai, Bo and Wu, Wayne and Loy, Chen Change},
  booktitle=ICCV,
  pages={13919--13929},
  year={2021}
}

@misc{basicsr,
  title={BasicSR: Open source image and video restoration toolbox},
  author={Wang, Xintao and Xie, Liangbin and Yu, Ke and Chan, Kelvin CK and Dong, Chao and Loy, Chen Change},
  howpublished={\url{https://github.com/XPixelGroup/BasicSR}},
  year={2022}
}

@inproceedings{rao2025rethinking,
  title={Rethinking video deblurring with wavelet-aware dynamic transformer and diffusion model},
  author={Rao, Chen and Li, Guangyuan and Lan, Zehua and Sun, Jiakai and Luan, Junsheng and Zhao, Lei and Lin, Huaizhong and Dong, Jianfeng and Xing, Wei},
  booktitle=ECCV,
  pages={421--437},
  year={2025},
  publisher={Springer}
}

@inproceedings{chen2018robust,
  title={Robust video content alignment and compensation for rain removal in a CNN framework},
  author={Chen, Jie and Tan, Cheen-Hau and Hou, Junhui and Chau, Lap-Pui and Li, He},
  booktitle=CVPR,
  pages={6286--6295},
  year={2018}
}
}

\end{document}